\documentclass[10pt,twocolumn,letterpaper]{article}

\usepackage{cvpr}
\usepackage{times}
\usepackage{graphicx}
\usepackage{amsmath}
\usepackage{amssymb}

\usepackage[american]{babel}
\usepackage{subfigure}
\usepackage{booktabs}
\usepackage{multirow}
\usepackage{pifont}
\newcommand{\cmark}{\ding{51}}%
\newcommand{\xmark}{\ding{55}}%

\usepackage[pagebackref=true,breaklinks=true,colorlinks,bookmarks=false]{hyperref}

\cvprfinalcopy 

\def\cvprPaperID{6503}

\begin{document}

\title{Neural Network Pruning with Residual-Connections and Limited-Data}

\author{Jian-Hao Luo \qquad Jianxin Wu\thanks{ This research was partially supported by the National Natural Science Foundation of China (61772256, 61921006). J. Wu is the corresponding author.}\\
	National Key Laboratory for Novel Software Technology\\Nanjing University, Nanjing, China\\
	{\tt\small luojh@lamda.nju.edu.cn, wujx2001@nju.edu.cn}
}

\maketitle

\begin{abstract}
	Filter level pruning is an effective method to accelerate the inference speed of deep CNN models. Although numerous pruning algorithms have been proposed, there are still two open issues. The first problem is how to prune residual connections. We propose to prune both channels inside and outside the residual connections via a KL-divergence based criterion. The second issue is pruning with limited data. We observe an interesting phenomenon: directly pruning on a small dataset is usually worse than fine-tuning a small model which is pruned or trained from scratch on the large dataset. Knowledge distillation is an effective approach to compensate for the weakness of limited data. However, the logits of a teacher model may be noisy. In order to avoid the influence of label noise, we propose a label refinement approach to solve this problem. Experiments have demonstrated the effectiveness of our method (CURL, Compression Using Residual-connections and Limited-data). CURL significantly outperforms previous state-of-the-art methods on ImageNet. More importantly, when pruning on small datasets, CURL achieves comparable or much better performance than fine-tuning a pretrained small model.   
\end{abstract}

\section{Introduction}

Deep neural networks have now become the dominating method in various computer vision fields, such as image recognition~\cite{resnet, AlexNet, VGG16} and object detection~\cite{rcnn}, and we have witnessed a great improvement in model accuracy. But, deploying a large CNN model on resource constrained devices like mobile phones is still challenging. Due to over-parameterization, it is both storage and time consuming to run a cumbersome large model on small devices. 

Network pruning is a useful tool to obtain a satisfactory balance between inference speed and model accuracy. Among these methods, filter level pruning aims to remove the whole unimportant filters according to a certain criterion. This strategy will not damage the original model structure and is attracting more and more attention recently.

\begin{figure}
	\centering
	\includegraphics[width=0.9\linewidth]{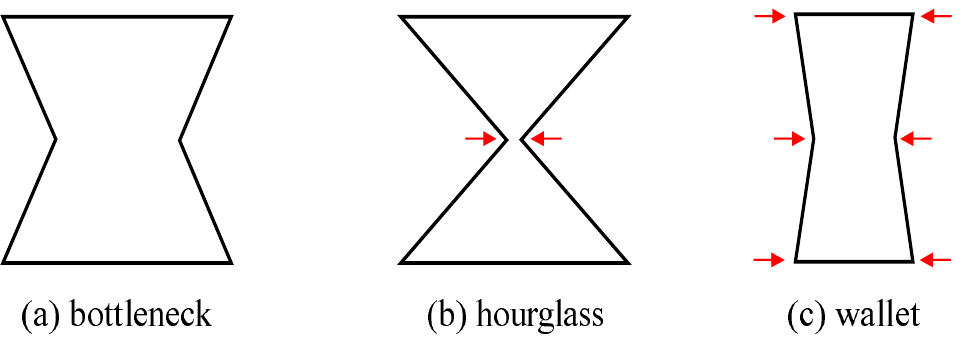}
	\caption{Illustration of residual block pruning with different strategies. (a) Bottleneck structure of residual blocks. (b) Only prune channels inside the bottleneck, generating an hourglass structure. (c) Prune channels both inside and outside the residual connection, generating a shape similar to an opened wallet.}
	\label{motivation}
\end{figure}

Although numerous filter level pruning algorithms have been proposed, there are still several open issues. First, \emph{pruning residual connections} is very difficult. As illustrated in Fig.~\ref{motivation}, most previous pruning methods only prune filters inside the residual connection, leaving the number of output channels unchanged. With a smaller target model (\ie, more filters pruned), the original bottleneck structure will become an hourglass. Obviously, representation ability of middle layers inside the hourglass structure is severely handicapped. Therefore, pruning channels both inside and outside the residual connection is more preferred for accelerating networks. Then, the pruned block is still bottleneck or in an opened wallet shape. 

As illustrated in the experiments section, the wallet structure has more advantages compared with hourglass: 1) it is more accurate thanks to a larger pruning space; 2) it is faster even with the same number of FLOPs; 3) it can save more storage space because more weights will be pruned.

The second issue is about \emph{pruning models with limited data}. Most current pruning methods only report their results on toy datasets (\eg, MNIST~\cite{MNIST}, CIFAR~\cite{CIFAR}) or large scale datasets (\eg, ImageNet~\cite{ImageNet}), ignoring an important real application scenario: pruning models on small datasets which have few images per category. This is a very common requirement, because we will not apply ImageNet in a real application. Directly pruning on a target dataset (which is usually small) is necessary.

In order to get a small model on a target small dataset, there are two different ways: 1) compress the network using the large dataset (or using a small network trained from scratch on the large dataset), and then fine-tune on the target small dataset; 2) directly prune the model \emph{without} access to the large dataset. In many real-world scenarios, the only choice is to compress the network using the small dataset and fine-tune on the same small dataset.

But, the reality is that \emph{directly pruning on a small dataset usually has a significantly lower accuracy than fine-tuning a small model which is pruned or trained from scratch on the large scale dataset}. This phenomenon widely exists in various networks and datasets. For example, as shown in ThiNet~\cite{luo2017thinet}, fine-tuning a pruned model which is compressed on ImageNet is a better choice when transferring to other domains. They found that the accuracy of directly pruning on CUB200~\cite{CUB200} is only 66.90\%, while fine-tuning a pruned ImageNet model can achieve 69.43\%. A dilemma is that directly pruning on the target dataset is often the case in real-world applications, where large datasets are either proprietary or too expensive to be used by ordinary users.

In this paper, we propose CURL, namely \textbf{C}ompression \textbf{U}sing \textbf{R}esidual-connections and \textbf{L}imited-data, to address both issues. In order to prune the channels outside of the residual connection, we show that all the blocks in the same stage should be pruned simultaneously due to the shortcut connection. We propose a KL-divergence based criterion to evaluate the importance of these filters. The channels inside and outside the residual connections will both be pruned, leading to a wallet shaped structure. Experiments on ImageNet show that the proposed residual block pruning method outperforms the previous state-of-the-art. To address the problem caused by the lack of enough training data, we propose to combine knowledge distillation~\cite{hinton2014distilling} and mixup~\cite{mixup} together and enlarge the training dataset via image transformation. We also propose a novel method to correct the noise in the logits of the teacher model. All the techniques greatly improve the accuracy of directly pruning with limited data.

Our contributions are summarized as follows.
\begin{itemize}
    \item We propose a novel way to compress residual blocks. We prune not only channels inside the residual branch, but also channels of its output activation maps (both the identity branch and the residual branch). The resulting wallet-shaped structure shows more advantages than previous hourglass-shaped structure. 
    
    \item Data augmentation is very effective in model fine-tuning with limited data. We show that combining data augmentation and knowledge distillation can achieve better performance. To avoid the influence of label noise, we propose a label refinement strategy which can further improve the accuracy.
    
\end{itemize}

\section{Related Work}
Pruning is an effective method to accelerate model inference speed and to reduce model size. Recent developments on network pruning can be roughly divided into two categories, non-structured and structured pruning. 

In the early stage, researchers mainly focused on non-structured pruning. Han \etal~\cite{han2, han1} proposed a magnitude-based pruning method to remove redundant weights. Connections with small absolute values are regarded as unimportant and are discarded. In order to compensate for the unexpected loss, Guo~\etal~\cite{guo2016dynamic} incorporated the splicing operation into network pruning. Once the pruned connections are found to be important, they could be recovered at any time.

However, the weakness of non-structured pruning is obvious. Due to the cache and memory access issues caused by irregular connections, its actual inference speed will be adversely affected. Therefore, structured pruning such as filter level pruning is more preferred.

In filter level pruning, the whole filter is discarded if it is considered to be unimportant. Importance evaluation criterion plays a crucial role in the success of pruning. Li~\etal~\cite{weight_sum} introduced previous magnitude based criterion into filter level pruning and calculated importance scores according to its $\ell_1$-norm. He~\etal~\cite{he2017channel} proposed to select filters by a LASSO regression based method and least square reconstruction. Luo~\etal~\cite{luo2017thinet} calculated filter importance based on statistics computed from its next layer. He~\etal~\cite{he2019filter} calculated the geometric median of the filters within one layer to prune filters with redundant information.

There are also some explorations without explicitly calculating the importance of each filter. Liu~\etal~\cite{liu2017learning} imposed $\ell_1$ regularization on the scaling factors of batch normalization layers to select unimportant channels. Huang and Wang~\cite{huang2018data} also introduced scaling factors into the model training process. By forcing some of the factors to zero, the unimportant filters will be pruned. Luo~\etal~\cite{luo2018autopruner} designed an efficient channel selection layer to find less important filters in an end-to-end manner. He~\etal~\cite{he2018amc} leveraged reinforcement learning to efficiently sample the design space and achieved better compression performance. Yu~\etal~\cite{slimmable} proposed slimmable neural networks, a general method to train a single CNN model executable at different widths. Although this approach is not designed for pruning, the output dimensions of residual blocks are reduced. But, except slimmable networks, the pruned models of the above filter pruning methods are all hourglass shaped.

To sum up, these pruning methods achieve pretty good results on toy datasets or large scale datasets and contribute a lot to the development of network pruning. But they all ignored the problem of pruning with limited data, and the pruned models are hourglass shaped. To the best of our knowledge, this is the first attempt to solve the problem of pruning networks on a small limited dataset.

\section{Our Method}
In this section, we will propose our method, CURL, which stands for \textbf{C}ompression \textbf{U}sing \textbf{R}esidual-connections and \textbf{L}imited-data. As its name implies, CURL consists of two major parts: prune residual-connections and prune with limited small-scale data. 

\subsection{Prune Residual-Connections}
The first step is to evaluate the importance of each filter, and prune away some less important filters to get a small model. We will give a quick recap of residual-connections, including its structure and the weakness of current evaluation criterion. Then, our new method will be presented.

\subsubsection{Recap of Residual-Connections}
\begin{figure}
	\centering
	\includegraphics[width=0.85\linewidth]{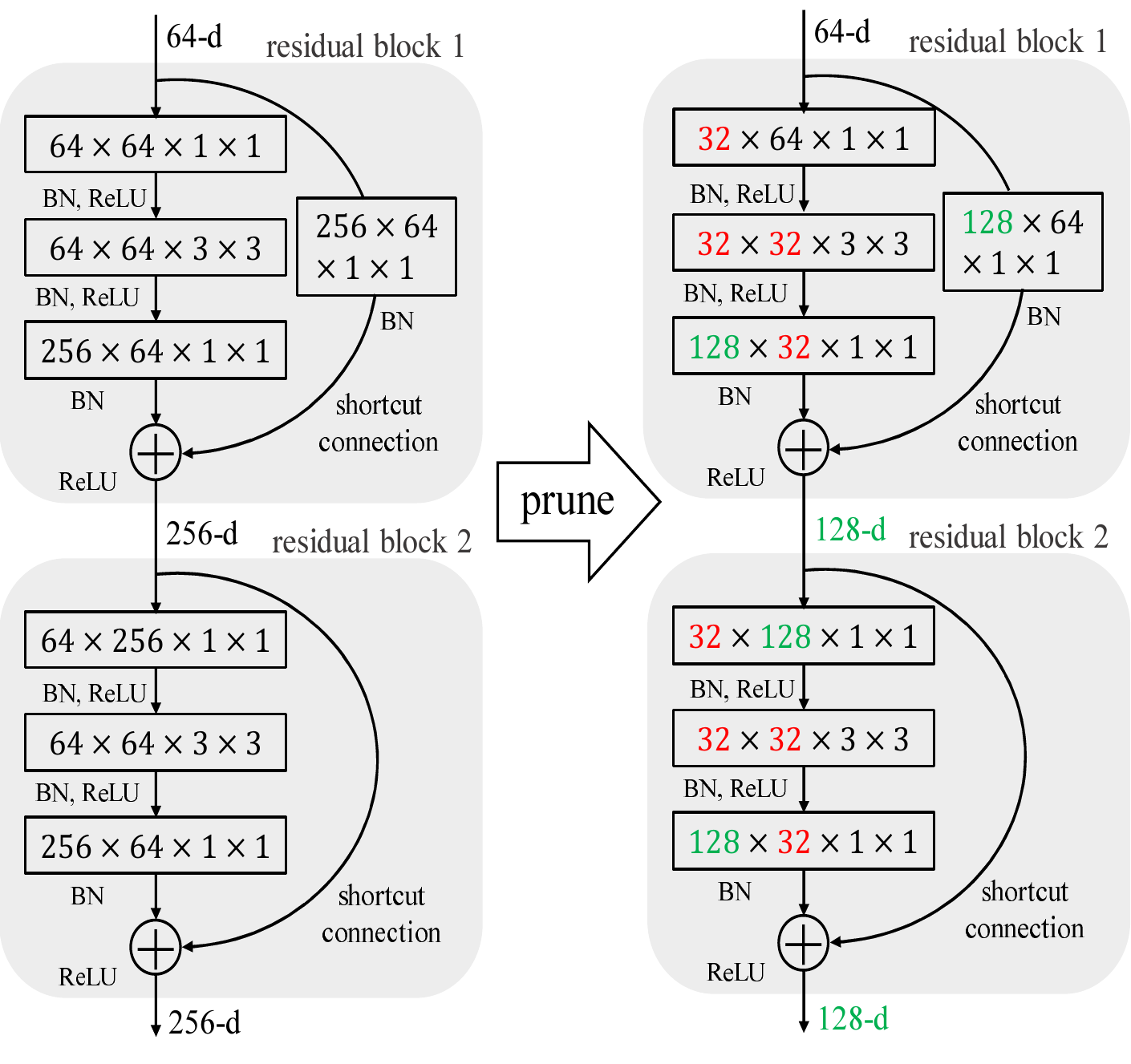}
	\caption{Illustration of residual block pruning strategy. Our method prunes not only channels inside the residual block (red numbers), but also channels of its output (green numbers). The first two numbers in each rectangle (convolutional layer) represent the number of output and input channels in a layer, respectively. This figure is best viewed in color.}
	\label{residual_pruning}
\end{figure}

Fig.~\ref{residual_pruning} illustrates our strategy for residual block pruning. The left part shows a typical residual structure of ResNet~\cite{resnet}. Suppose there are two residual blocks in the current stage. Each block consists of three convolutional layers (including the batch normalization layer and the ReLU activation layer). Usually, a down-sample layer is necessary to process the different activation sizes and channel numbers between two stages.

Due to the existence of the shortcut connection, the channel numbers of all residual blocks in a stage need to be consistent in order for the sum operations to be valid. Hence, pruning residual connection is very difficult. Most previous studies only focus on reducing channels inside the residual block (as shown in the red numbers of Fig.~\ref{residual_pruning}), leaving the output dimension unchanged. 

However, there is a great need for residual connection's pruning. Firstly, pruning both inside and outside channels is faster than only pruning inside channels during inference, even though their FLOPs are the same. Secondly, since the activation map of each block is reduced, we can save more memory spaces. Last but not least, reducing residual output means our pruning space is enlarged. We can achieve higher accuracy with the same compression ratio.

In order to reduce the output dimension, all the blocks in the same stage should be pruned simultaneously (as well as the down-sample layer). This is not a simple task, which can not be finished with previous method. Most of the previous importance evaluation criterion only focus on single layer (\eg, the ThiNet method~\cite{luo2017thinet}), and ignore the relation of other layers. Hence, we should design a new importance criterion that can evaluate multiple filters simultaneously. 

\subsubsection{The Proposed Pruning Method}
Inspired by ThiNet~\cite{luo2017thinet}, we propose a new method that can evaluate filter importance globally. The main idea of ThiNet is to minimize the reconstruction error of the next layer's output. Similarly, our goal is to minimize the information loss of the last layer (\ie, softmax layer).

Let us take Fig.~\ref{residual_pruning} as an example. The output dimension of residual blocks is 256. Now we want to evaluate the importance score of each channel. A natural idea is to remove the output channels one by one, and calculate the information loss after channel removal. Due to the structure constraint, the output channels of each residual block should be removed simultaneously. For example, the first output channel of block 1, block 2 \emph{and} the down-sample layer should be removed simultaneously in Fig.~\ref{residual_pruning}.

Inspired by network slimming~\cite{liu2017learning}, we will reset the parameters of the BN layers to remove the corresponding filters. The output channel of each block is calculated by
\begin{equation}
	y_i = \gamma \frac{x_i-\mu_\mathcal{B}}{\sqrt{\sigma^2_{\mathcal{B}}+\epsilon}} + \beta,
\end{equation}
where $\gamma, \beta, \mu_\mathcal{B}, \sigma_{\mathcal{B}}$ are the batch-normalization~\cite{BN} parameters. Since BN is channel independent, we can simply set $\gamma=\beta=0$ and the corresponding output channel will be zeroed out. It means the corresponding filters of each output block are removed.

Then, we will evaluate the performance change before/after channel removal. Hence, we need a proxy dataset. Naturally, the proxy dataset could be the training dataset. However, using all the training images can be messy and time-consuming. We randomly select 256 images from the training dataset, and extract the prediction probability (the output of the softmax layer) on these 256 images.

Let $p$ be the output probability of the original network, and $q$ be the probability after channel removal. A popular method to compare the similarity of two probability distributions is KL-divergence. We calculate the filter importance score as
\begin{equation}\label{KL_criterion}
	s = \text{D}_\text{KL}(p||q)=\sum_{i=1}^n p_i\text{log}\frac{p_i}{q_i}.
\end{equation}
If the current filter is redundant, $s$ will approach 0. Removing this filter has almost no influence to the prediction results. Conversely, a larger value of $s$ means the current dimension is more important. Hence, using the KL-divergence to denote the importance score is reasonable. It can reflect the information loss of removing some filters. This step will be repeated 256 times, resulting in 256 importance scores, one for each channel.

As for those channels inside the residual block (red numbers of Fig.~\ref{residual_pruning}), the process is easier. We only need to erase one filter of the current layer at each step. Finally, filter scores in all layers will be sorted in the ascending order. The top $k$ filters will be removed, leading to a pruned small model. In practice, the value of $k$ depends on the available computational or storage budget. To prevent any extreme issues (\eg, there are only few filters left in a layer after pruning), the smallest compression rate of each layer should not be smaller than a threshold (\eg, 0.3). 

The benefits of our pruning method are obvious. First, it is a global criterion which can evaluate the impact of all filters simultaneously. Such kind of global criterion makes possible the pruning of residual connections. Secondly, there is no direct correlation between importance score and filter location. In previous magnitude-based methods (\eg, $\ell_1$-norm of each filter~\cite{weight_sum}), the magnitude of importance scores among layers are different. Hence, a fixed compression ratio should be specified for each layer. However, we can achieve an adaptive compression. How many filters should be removed entirely depends on the scores.

\subsection{Prune with Limited Data} \label{sec_fine_tuning}
The pruned small model is then fine-tuned on the target dataset. Fine-tuning with limited small-scale data is challenging. Data-augmentation (\eg, mixup~\cite{mixup}) plays an important role in this step. As aforementioned, ImageNet pretraining is a crucial augmentation method which achieves much higher accuracy than directly pruning on the small dataset. The major difference between ImageNet and small dataset is the number of training examples. In order to compensate for this gap, we use several image transformation techniques to expand the training dataset and fine-tune the pruned small model with knowledge distillation~\cite{hinton2014distilling}. However, the logits may be noisy (since the teacher model has not seen these augmented data). We then propose a label refinement method to update these noisy logits.

\subsubsection{Data Expansion}

\begin{figure} 
	\centering    
	\subfigure[the original image] {
		\label{fig:a}     
		\includegraphics[width=0.3\columnwidth]{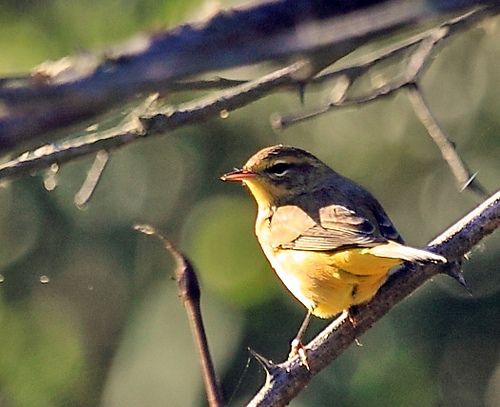}  
	}     
	\subfigure[rotate] { 
		\label{fig:b}     
		\includegraphics[width=0.3\columnwidth]{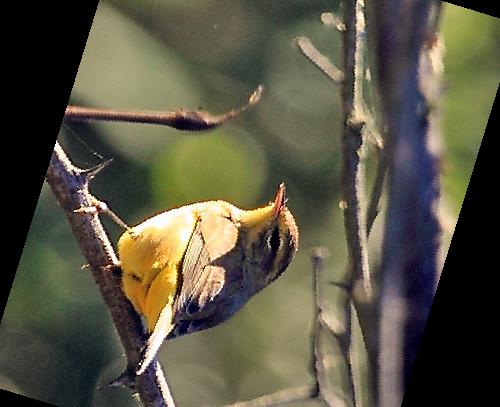}  
	}    
	\subfigure[cutout~\cite{cutout}] { 
		\label{fig:c}     
		\includegraphics[width=0.3\columnwidth]{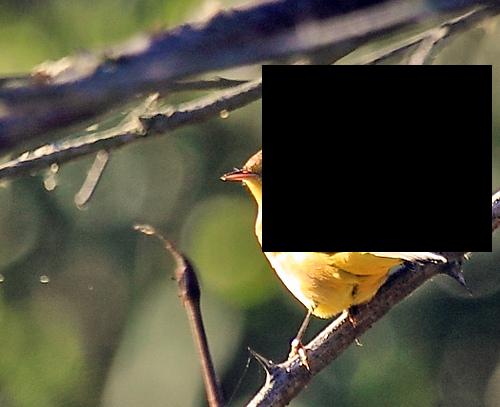} 
	}
	\subfigure[$2\times 2$ shuffle~\cite{chen2019destruction}] {
		\label{fig:d}     
		\includegraphics[width=0.3\columnwidth]{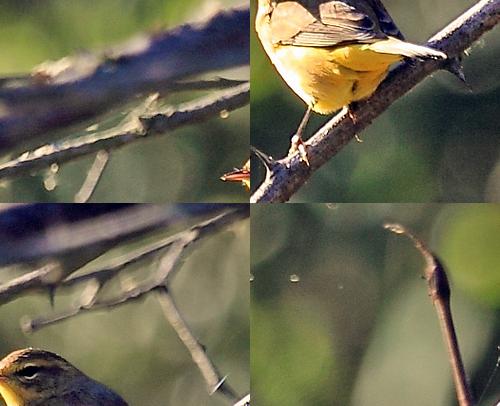}  
	}     
	\subfigure[$3\times 3$ shuffle~\cite{chen2019destruction}] { 
		\label{fig:e}     
		\includegraphics[width=0.3\columnwidth]{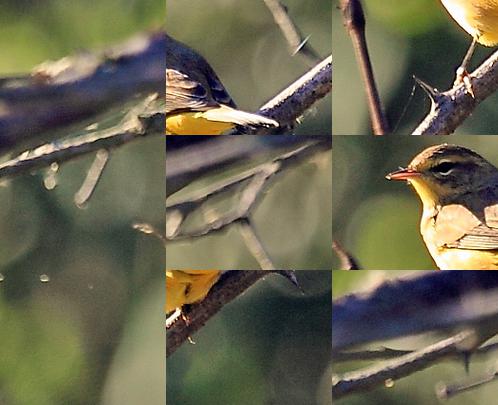} 
	}    
	\subfigure[$4\times 4$ shuffle~\cite{chen2019destruction}] { 
		\label{fig:f}     
		\includegraphics[width=0.3\columnwidth]{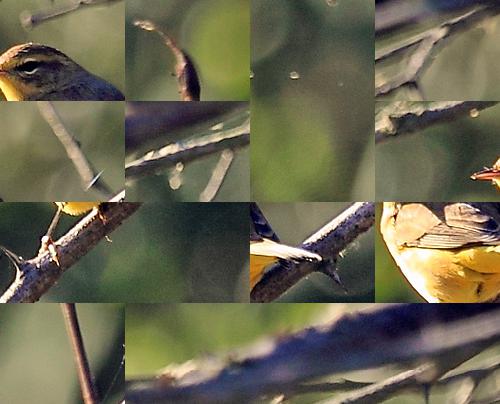} 
	}   
	\caption{Visualization of the expanded dataset.}     
	\label{dataset}     
\end{figure}

Since the main difficulty of fine-tuning on small dataset is caused by limited data, a natural idea is to generate or collect more training images. For example, Chen~\etal~\cite{Chen_2019_ICCV} exploited GAN to generate training samples. However, training the generator network can be a more challenging issue, especially for real images with large resolution. Here, we adopt a simpler method to expand the training dataset.

Fig.~\ref{dataset} shows examples of data expansion results. We use three different image transformation techniques:
\begin{itemize}
	\item Rotate: randomly rotate the original images $r$ degrees, where $r \in [0, 360)$.
	\item Cutout~\cite{cutout}: cut out a rectangle patch from the original image pixels with random location. The side length of rectangle is randomly selected from $[0.2, 0.5]$ times the original image size.
	\item Shuffle~\cite{chen2019destruction}: uniformly partition the image into $N\times N$ sub-regions, and shuffle these partitioned local regions. We use three different partition sizes, namely $2\times2$, $3\times3$, $4\times4$.
\end{itemize} 

Our motivation behind these transformation is that the most discriminative information often lies in local image patches (\eg, leg color of a bird) for object recognition problems. In order to find these discriminative regions, the network should pay more attention to local patches rather than global information. Hence, we can cut out some regions or even shuffle the whole image to remove the influence of global information (\ie, background). 

\subsubsection{Label Refinement}
We use knowledge distillation~\cite{hinton2014distilling} to fine-tune the small model. The original large model plays the role of a teacher model and the pruned small model is the student model. A typical strategy is to train the student model under the supervision of soft target (the logits of teacher models) and hard target (the groundtruth label). However, because the teacher model has not seen the new data, its output (logits) may be noisy.

Inspired by PENCIL~\cite{pencil} and R2-D2~\cite{R2D2}, we can update the noisy logits during model training via SGD. However, updating logits is dangerous. The quality of soft targets can be updated to become worse if the student model is not accurate enough. To avoid this situation, we divide the whole training process into two steps: 
fine-tuning on \emph{original small dataset} with knowledge distillation plus mixup and fine-tuning on \emph{expanded dataset} with label refinement. Note that the expanded new data and our proposed refinement method are only used in step 2.

\textbf{Step 1: knowledge distillation with mixup.} Knowledge distillation~\cite{hinton2014distilling} and mixup~\cite{mixup} are two widely used techniques, which are really helpful for the training with limited data. We propose to combine these two techniques together. First, a new input is generated via mixup:
\begin{equation}
	\tilde{x} = \lambda x_i + (1-\lambda)x_j, \quad \tilde{y} = \lambda y_i + (1-\lambda)y_j,
\end{equation}
where $(x_i, y_i)$ and $(x_j, y_j)$ are two random examples, $\lambda \in [0, 1]$ is drawn from a Beta distribution. The new input $\tilde{x}$ is then fed into teacher and student models. Let $u$ denote the output logits of the teacher model, and $v$ denote the student's output. Our total loss is calculated as:
\begin{equation}
	\mathcal{L}=\alpha T^2 \mathcal{L}_\text{KL}(p, q) + (1-\alpha) \mathcal{L}_\text{CE}(q, \tilde{y}),
	\label{eq_step1}
\end{equation}
where $p=\text{softmax}(u/T)$, $q=\text{softmax}(v/T)$ are the softmax output under temperature $T$. $\mathcal{L}_\text{KL}$ and $\mathcal{L}_\text{CE}$ denote KL-divergence loss and cross-entropy loss, respectively. $\alpha \in [0, 1]$ controls the balance between losses. With these two techniques, the pruned small model can converge into a good local minima.

\begin{figure}
	\centering
	\includegraphics[width=0.9\linewidth]{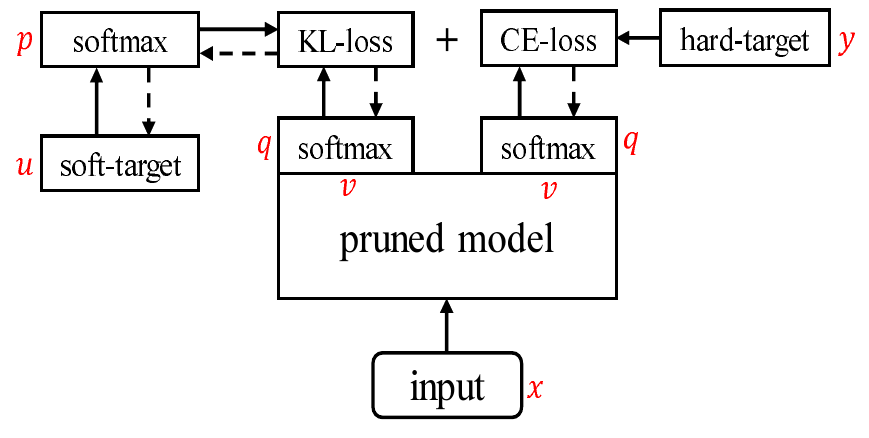}
	\caption{Illustration of label refinement. The soft target (teacher model's logits) will also be updated during model fine-tuning. Solid and dashed arrows denote forward and backward propagation, respectively.}
	\label{refinement}
\end{figure}
\textbf{Step 2: knowledge distillation with label refinement.} We then fine-tune the small model on our expanded dataset and update logits to remove label noises. Fig.~\ref{refinement} illustrates the framework of our label refinement process. The soft target (\ie, teacher model's logits) of each image in our expanded dataset will be extracted first, and stored in memory. Our loss function is designed as:
\begin{equation}
\mathcal{L}=\alpha T^2 \mathcal{L}_\text{KL}(q, p) + (1-\alpha) \mathcal{L}_\text{CE}(q, y).
\end{equation}
The notations are the same as Eq.~(\ref{eq_step1}). Note that mixup is not used here. The major difference with knowledge distillation is that we use a reversed KL-divergence loss to encourage the model to pay more attention to label refinement, as in~\cite{R2D2, pencil}. During model fine-tuning, the soft-target will also be updated via SGD:
\begin{equation}
	u \longleftarrow u - \eta \cdot \nabla_u\mathcal{L} \,,
\end{equation}
where $\eta$ is the learning rate for updating $u$, $\nabla_u\mathcal{L}$ denotes the gradient of the loss function $\mathcal{L}$ with respected to $u$ which is calculated via back-propagation.

\section{Experiments}
In this section, we will evaluate the performance of CURL. In order to compare CURL with state-of-the-art methods, we first test the effectiveness on ImageNet~\cite{ImageNet}. ResNet50~\cite{resnet} will be pruned on this dataset. Then, more results on small-scale datasets will be presented. Our method achieves better or comparable performance on these datasets than small models with ImageNet pretraining. Finally, we will end this section with ablation studies. All the experiments were conducted with PyTorch~\cite{pytorch}.

\subsection{Pruning ResNet50 on ImageNet} \label{exp_ImageNet}
\textbf{Implementation details.} We follow the previous training settings. For a fair comparison, all the fine-tuning techniques introduced in Section~\ref{sec_fine_tuning} are \emph{not used} here. In other words, we do not use mixup, knowledge distillation or dataset expansion techniques in the ImageNet experiment. Once the filter importance has been evaluated by CURL, we will remove all unimportant filters of all layers. The pruned model is then fine-tuned with 100 epochs. Data argumentation strategy and parameter settings are the same as PyTorch official examples. We adopt a large mini-batch size 512. The initial learning rate is set to $0.1$. Warmup~\cite{warmup} and cosine learning rate decay~\cite{cosine_decay} are also used here. Since the output of the last stage is closely related to prediction, we will not prune the output dimension of the last residual stage (\ie, the output channel number is still 2048).

\begin{table}
	\caption{Comparison results of pruning ResNet50 on ImageNet.}
	\small
	\label{resnet_imagenet}
	\setlength{\tabcolsep}{3pt} 
	\centering
	\begin{tabular}{lcccr}
		\toprule
		Method & Top-1 Acc. & Top-5 Acc.& MACs & \#Param.\\
		\midrule
		ResNet50 & 76.15\% & 92.87\%& 4.09G & 25.56M\\
		ThiNet-30~\cite{luo2017thinet} & 68.42\% & 88.30\% & 1.10G & 8.66M \\
		GAL-1-joint~\cite{GAL} & 69.31\% & 89.12\% & 1.11G & 10.21M \\
		GDP-0.5~\cite{GDP} & 69.58\% & 90.14\% & 1.57G & - \\
		Taylor-FO~\cite{molchanov2019importance} & 71.69\% & - & 1.34G & 7.90M \\
		Slimmable NN~\cite{slimmable} & 72.10\% & 90.57\% & 1.05G & 6.92M \\
		AutoPruner~\cite{luo2018autopruner} & 73.05\% & 91.25\% & 1.39G & 12.60M\\
		CURL & \textbf{73.39\%} & \textbf{91.46\%} & 1.11G & 6.67M \\
		\bottomrule
	\end{tabular}
\end{table}

Table~\ref{resnet_imagenet} shows the pruning results of ResNet50. We test model accuracy using 1-crop validation: the shorter side is resized to 256, followed by a $224\times 224$ center crop and mean-std normalization. The accuracy of the last epoch will be reported. Obviously, CURL is significantly better than previous state-of-the-art. Our method obtains a higher accuracy even with fewer MACs (Multiply-ACcumulate operations) and parameters.

In order to demonstrate our motivation for residual connection pruning, we also test the actual inference speed on a NVIDIA Tesla M40 GPU. AutoPruner~\cite{luo2018autopruner} only prunes channels inside the residual block. With a small compression rate (more filters will be removed), the middle layer of a residual block will be very thin, leading the bottleneck structure into hourglass. By contrast, we prune both channels inside and outside residual blocks. The pruned structure is still bottleneck or opened wallet. 

Wallet is not only more accurate than hourglass (which has been demonstrated above), but also faster even with the same number of MACs. For a fair comparison, we adjust the threshold to obtain a new model with 1.39G MACs and 7.83M parameters. The inference time of hourglass on M40 GPU to process a mini-batch of 256 images is 0.21s, while wallet only cost 0.19s. More importantly, the model size of wallet is much smaller than hourglass (nearly halved). Hence, memory consumption could be greatly reduced with our pruned structure.

\subsection{Pruning on Small-Scale Datasets}
We then prune large models on small datasets. Our goal is to demonstrate that directly pruning with limited data using CURL could achieve comparable or even better performance than those small models pruned or trained with ImageNet. Two widely used network, namely ResNet50~\cite{resnet} and MobileNetV2~\cite{mobilenetv2}, will be pruned on four small-scale datasets. Table~\ref{tbl:dataset} summarizes the information of these four datasets, including training and validation sizes.

\begin{table}
	\caption{Summary of 4 small datasets. }
	\small
	\label{tbl:dataset}
	\setlength{\tabcolsep}{3pt} 
	\centering
	\begin{tabular}{lcrrr}
		\toprule
		Dataset Name & Meta-Class & \#Train& \#Val. & \#Categories\\
		\midrule
		CUB200-2011~\cite{CUB200} & Birds & 5994 & 5794 & 200 \\
		Oxford Pets~\cite{pets} & Cats\&Dogs & 3680 & 3669 & 37 \\
		Stanford Dogs~\cite{Dogs} & Dogs & 12000 & 8580 & 120 \\
		Stanford Car~\cite{Car} & Cars & 8144 & 8041 & 196 \\
		\bottomrule
	\end{tabular}
\end{table}

\begin{table*}
	\caption{Pruning results on small-scale datasets. In each model, there are 4 methods: 1) fine-tune the large model which is trained on ImageNet; 2) fine-tune the small model which is trained or pruned on ImageNet; 3) directly prune the large model on small dataset using slimmable neural network~\cite{slimmable}; 4) directly prune the large model on small dataset using CURL.}
	\small
	\label{small_dataset}
	\setlength{\tabcolsep}{6pt} 
	\centering
	\begin{tabular}{clcccccccc}
		\toprule
		\multirow{2}{*}{Model} & \multirow{2}{*}{Method} & \multicolumn{2}{c}{CUB200-2011~\cite{CUB200}} & \multicolumn{2}{c}{Pets~\cite{pets}} & \multicolumn{2}{c}{Dogs~\cite{Dogs}} & \multicolumn{2}{c}{Car~\cite{Car}}\\
		\cline{3-10} & & Acc. & MACs  & Acc. & MACs  & Acc. & MACs  & Acc. & MACs  \\
		\midrule
		\multirow{4}{*}{MobileNetV2} & MobileNetV2-1.0 & 78.77\% & 299.77M & 89.94\% & 299.56M & 78.94\% & 299.67M & 87.27\% & 299.76M \\
		& MobileNetV2-0.5 & 73.96\% & 96.12M & 86.32\% & 95.91M & 72.15\% & 96.02M & 81.50\% & 96.11M \\
		& Slimmable NN~\cite{slimmable} & 72.20\% & 96.12M & 83.37\% & 95.91M & 67.81\% & 96.02M & 87.02\% & 96.11M \\
		& CURL & \textbf{78.72\%} & 96.07M & \textbf{86.89\%} & 95.91M & \textbf{74.72\%} & 96.04M & \textbf{87.64\%} & 96.15M \\
		\midrule
		\multirow{4}{*}{ResNet50} & ResNet50 & 84.76\% & 4.09G & 92.59\% & 4.09G & 83.82\% & 4.09G & 91.89\% & 4.09G \\
		& ResNet50-CURL & 81.33\% & 1.11G & \textbf{90.84\%} & 1.11G & \textbf{81.60\%} & 1.11G & 88.60\% & 1.11G \\
		& Slimmable NN~\cite{slimmable} & 80.05\% & 1.05G & 89.67\% & 1.05G & 77.80\% & 1.05G & 91.47\% & 1.05G\\
		& CURL & \textbf{83.64\%} & 1.10G & 90.30\% & 1.11G & 79.79\% & 1.11G & \textbf{92.19\%} & 1.11G \\
		\bottomrule
	\end{tabular}
\end{table*}

\textbf{Baseline setting.} We adopt three baselines for comparison: 1) Fine-tune the large model which is pretrained on ImageNet. This is a typical approach to get a classification network on a target task. 2) Fine-tune a small model which is pretrained (MobileNetV2-0.5) or pruned (ResNet50-CURL, the CURL model in Table~\ref{resnet_imagenet}) on ImageNet. This is a compromise due to the difficulty of pruning with limited data. 3) Directly prune the fine-tuned large model (\ie, baseline 1) on the target small dataset using the slimmable neural network method~\cite{slimmable}. During fine-tuning, the models are trained with the mini-batch size of 32 for 300 epochs. Learning rate is initialized as 0.001, and reduced with cosine schedule. Warmup is used in the first 5 epochs. Mixup is also adopted with $\alpha=1$. Other parameter settings and data argumentation strategies are the same as PyTorch official examples. 

\textbf{Implementation details of CURL.} The fine-tuning process of CURL is divided into two steps. For a fair comparison, the total fine-tuning epochs are still 300. In the first step, the pruned model is trained with mixup and knowledge distillation in 200 epochs. The temperature $T$ of knowledge distillation is set to 2, $\alpha$ in Eq.~(\ref{eq_step1}) is set to 0.7, and learning rate is 0.01. Warmup and cosine decay schedule are also used. In the second step, since the dataset is enlarged 6 times, we train the small model with 16 epochs ($16\times6 \approx 100$). Because logits will be updated during fine-tuning, the model is not sensitive to temperature, we set $T=1$ here. The value of $\alpha$ is not changed. The learning rate for pruned network is set to 0.0001 and reduced using cosine schedule. As for the learning rate $\eta$ for updating logits, we set it to 1 and will not change during training. Other parameter settings are the same as those used in baselines. The pruning strategy for ResNet is the same as experiment on ImageNet. As for MobileNetV2, all the bottleneck will be pruned. Note that the second layer of each bottleneck is depth-wise convolution, which means the input dimension should be the same as its output. Hence, the first two layers of each residual block will be pruned simultaneously.

Table~\ref{small_dataset} shows results on these four datasets. Performance comparison between the second and third lines prove that \emph{with limited training data, directly pruning on a small dataset is usually worse than fine-tuning a small model trained or pruned on ImageNet.} In other words, ImageNet pretraining is still a strong technique to get a more accurate model with limited data. However, this technique may not be applicable in many real-world applications. Training on large scale datasets is cumbersome and time-consuming, too. The original large dataset may be even not available in many real-world scenarios.

In stark contrast, the proposed CURL approach works well on most datasets. Our method achieves comparable or even better performance than fine-tuning small models trained or pruned on ImageNet. When MobileNetV2-1.0 model is pruned, CURL outperforms MobileNetV2-0.5 on all datasets by a large margin. It is worth mentioning that on Oxford Car, CURL is even better than the original large model. This result demonstrates that the proposed CURL method is really useful when prune with limited data. 

As for ResNet, since ResNet50-CURL is a very strong baseline (ResNet50-CURL achieves the best performance on ImageNet compared with previous state-of-the-art \emph{and it is already trained using CURL}), the advantage of directly pruning on small dataset is not so significant. However, we can still achieve better results than ResNet50-CURL on CUB200-2011 and Oxford Car. 

\subsection{Ablation Studies}
To explore the impact of different modules of CURL, we will perform an ablative study in this section. Three major modules, namely fine-tuning strategy, pruning criterion and label refinement will be studied. All the ablation studies are conducted on CUB200-2011~\cite{CUB200} with the MobileNetV2~\cite{mobilenetv2} model.
\subsubsection{Impact of Fine-Tuning Strategy}

\begin{table}
	\caption{Pruning MobileNetV2 on CUB200 with different fine-tuning methods. Here, ``nothing'' indicates a standard fine-tuning process, ``KD'' is knowledge distillation, and ``scratch'' denotes training from scratch.}
	\small
	\label{ablation_1}
	\setlength{\tabcolsep}{2pt} 
	\centering
	\begin{tabular}{ccccccc}
		\toprule
		Method & nothing & +mixup & +KD & +mixup\&KD & ours & scratch\\
		\midrule
		Acc. (\%) & 70.76 & 73.89 & 77.87 & 77.99 & 78.72 & 64.39 \\
		\bottomrule
	\end{tabular}
\end{table}

We first study the impact of different fine-tuning strategies. The large fine-tuned model (MobileNetV2-1.0) is evaluated and pruned by CURL, generating a small but not accurate model. Then the pruned model is fine-tuned using several different strategies with 300 epochs. Other parameter settings are the same as in previous experiments.

Table~\ref{ablation_1} summarizes the results of different fine-tuning methods. A standard approach is to train the small model without augmentation techniques like mixup or knowledge distillation. Unfortunately, the final accuracy is only 70.76\%, which is much worse than MobileNetV2-0.5 fine-tuned by the same approach (72.97\%). If mixup is used, the accuracy can be improved to 73.89\%, but still worse than MobileNetV2-0.5 fine-tuned with mixup (73.96\%). Note that if we prune the model using Slimmable NN~\cite{slimmable} with the same fine-tuning strategy, its accuracy is only 72.20\% (see Table~\ref{small_dataset} for more details). This phenomenon suggests that our KL-divergence based criterion, \ie, Eq.~(\ref{KL_criterion}), works well in evaluating filter importance. On the other hand, these results also demonstrate our motivation: directly pruning on a small dataset is usually worse than fine-tuning a small model which is trained or pruned on the large dataset. Although our pruning method outperforms previous state-of-the-art on ImageNet, fine-tuning on small dataset is still very challenging due to the lack of enough training data.

Knowledge distillation contributes a lot to the success of pruning with limited data as shown in~\cite{bai2019few}. If we equip the standard fine-tuning approach with knowledge distillation, the final accuracy can be improved to 77.87\%. If we combine mixup and knowledge distillation together, the accuracy can be further improved to 77.99\%. Based on these observations, we propose to enlarge the training dataset with image transformation methods, and update teacher model's logits with label refinement. Finally, the pruned small model can reach an accuracy of 78.72\%, which is almost the same as unpruned large model (78.77\%). Please note that the combination of mixup and knowledge distillation, and the updating of teacher's logits are both novel approaches proposed in our CURL in the network compression field.

The last experiment is about training from scratch. We randomly initialize the weights of pruned network structure and train the model by 1000 epochs with mixup. Starting from 0.05, the value of learning rate is also reduced by cosine schedule. Other parameter settings are the same as previous ones. However, the accuracy of training from scratch is only 64.39\%, much lower than any pruned results.

This phenomenon is reasonable, because training with limited data is very challenging. Recently, a study~\cite{liu2018rethinking} suggests that training the same network from scratch could achieve comparable or better performance than pruning. They think the real value of pruning is to find a better network structure rather than important filters. This conclusion may be correct with enough training time and data. However, when the conditions are not applicable (\eg, training with limited data), this conclusion will fail to hold true. In this situation, finding important filters becomes more valuable. To sum up, our pruning method provides a feasible solution for how to obtain a small model with limited data and how to get rid of the dependence on the large dataset, which is really important in many real-world applications. 

\subsubsection{Impact of Pruning Criterion}
\begin{table}
	\caption{Pruning MobileNetV2 on CUB200 with different evaluation criterion. We prune 50\% filters of the middle layers of the last bottleneck. The accuracy is tested without fine-tuning.}
	\small
	\label{ablation_2}
	\setlength{\tabcolsep}{5pt} 
	\centering
	\begin{tabular}{ccccc}
		\toprule
		Method & Random & Weight Sum & $\Delta$Loss & KL-Div. (Ours)\\
		\midrule
		Acc. & 6.80\% & 8.25\%  & 21.83\% & 66.33\% \\
		\bottomrule
	\end{tabular}
\end{table}

We then study the impact of different criterion for evaluating filter importance, \ie, Eq.~(\ref{KL_criterion}). We focus on the last bottleneck structure of MobileNetV2. There are two layers (point-wise convolution and depth-wise convolution) in this block. Due to the structure constraint of depth-wise convolution, we should prune these two layers simultaneously. 

The easiest way to finish this task is to randomly discard channels. We randomly prune 50\% filters of these two layers and get 6.80\% top-1 accuracy on CUB200  without fine-tuning (as shown in Table~\ref{ablation_2}). Starting from this simplest baseline, we then focus on how to design a better way to evaluate the importance of multiple filters. 

One possible solution is to extend previous magnitude-based criterion. Weight sum~\cite{weight_sum} is such kind of a method which calculates filter importance based on its $\ell_1$-norm. Here, we simply add the scores of two filters together: $s_i=\|W_i^1\|_1+\|W_i^2\|_1$, where $W_i^1$ and $W_i^2$ denote filter $i$ of the first (point-wise convolution) and second (depth-wise convolution) hidden layers, respectively. However, the extended version of weight sum is only a little better than the random strategy (8.25\% vs. 6.80\%).

We then consider data-driven criteria and concentrate on the output of the network. A natural idea is evaluating the change of accuracy. If removing some filters has no influence to the model accuracy, then we can regard them as unimportant. However, this approach is impractical. We find that model accuracy is not sensitive to one or two filters. Most channels will get the same score after evaluation. Since accuracy is closely related to loss, we can replace accuracy change with loss change and calculate the score as $s_i=\mathcal{L}_\text{pruned}-\mathcal{L}_\text{unpruned}$. The performance of this criterion is 21.83\%. Although this result is significantly better than random and weight sum, accuracy of the pruned network is still not satisfying.

Inspired by knowledge distillation, we propose a KL-divergence based criterion to evaluate the information loss of the pruned model. With this criterion, the accuracy is dramatically improved to 66.33\%.

\subsubsection{Impact of Label Refinement}

\begin{table}
	\caption{Pruning results on several small datasets with/without the proposed label refinement method.}
	\small
	\label{ablation_3}
	\setlength{\tabcolsep}{2pt} 
	\centering
	\begin{tabular}{cccccc}
		\toprule
		Model & Refinement? & CUB200 & Pets & Dogs & Car\\
		\midrule
		\multirow{2}{*}{MobileNetV2} & \xmark & 77.43\% & 86.65\% & 74.02\% & 87.15\% \\
		 & \cmark & 78.72\% & 86.89\% & 74.72\% & 87.64\% \\
		 \midrule
		 \multirow{2}{*}{ResNet50} & \xmark & 82.88\% & 90.32\% & 78.65\% & 91.77\% \\
		 & \cmark & 83.64\% & 90.30\% & 79.79\% & 92.19\% \\
		\bottomrule
	\end{tabular}
\end{table}

In order to compensate for the gap of limited data, we propose a label refinement technique. The original dataset is enlarged by 6 times via several image transformation methods. The fine-tuned model is then trained on this enlarged dataset with knowledge distillation. Due to label noise, soft target will also be updated during training.

Table~\ref{ablation_3} illustrates the effectiveness of our label refinement strategy. Compared with standard knowledge distillation equipped with mixup, using our label refinement method can achieve better performance on most datasets. 

Updating soft target can further improve model accuracy. For examples, if we set $\eta=0$ (do not update soft target) and fine-tune the pruned MobileNetV2 on the expanded CUB200 dataset, the accuracy is 78.56\%, slightly lower than the result of updating soft target (78.72\%). In this work, we adopt a simple method to process label noise of soft target. How to correct the misleading wrong labels is still worth exploring. If a more advanced algorithm is used, the model accuracy can be further improved.

\section{Conclusion}
In this paper, we proposed a novel filter level pruning method to accelerate the inference speed of deep CNN models. Different from previous pruning strategies, we prune both channels inside and outside the residual connections via a KL-divergence based criterion. We also propose a label refinement approach to avoid the influence of label noise. With the proposed CURL method, we can directly prune models on the small dataset and achieve comparable or even better results than fine-tuning a small model pruned or trained on the large scale dataset, which is of great value in many real-world scenarios.

\clearpage
{\small
\bibliographystyle{ieee_fullname}
\bibliography{egbib}

\begin{thebibliography}{10}\itemsep=-1pt

\bibitem{bai2019few}
Haoli Bai, Jiaxiang Wu, Irwin King, and Michael Lyu.
\newblock Few shot network compression via cross distillation.
\newblock {\em arXiv preprint arXiv:1911.09450}, 2019.

\bibitem{Chen_2019_ICCV}
Hanting Chen, Yunhe Wang, Chang Xu, Zhaohui Yang, Chuanjian Liu, Boxin Shi,
  Chunjing Xu, Chao Xu, and Qi Tian.
\newblock Data-free learning of student networks.
\newblock In {\em ICCV}, pages 3514--3522, 2019.

\bibitem{chen2019destruction}
Yue Chen, Yalong Bai, Wei Zhang, and Tao Mei.
\newblock Destruction and construction learning for fine-grained image
  recognition.
\newblock In {\em CVPR}, pages 5157--5166, 2019.

\bibitem{cutout}
Terrance DeVries and Graham~W Taylor.
\newblock Improved regularization of convolutional neural networks with cutout.
\newblock {\em arXiv preprint arXiv:1708.04552}, 2017.

\bibitem{rcnn}
Ross Girshick, Jeff Donahue, Trevor Darrell, and Jitendra Malik.
\newblock Rich feature hierarchies for accurate object detection and semantic
  segmentation.
\newblock In {\em CVPR}, pages 580--587, 2014.

\bibitem{warmup}
Priya Goyal, Piotr Doll{\'a}r, Ross Girshick, Pieter Noordhuis, Lukasz
  Wesolowski, Aapo Kyrola, Andrew Tulloch, Yangqing Jia, and Kaiming He.
\newblock {Accurate, large minibatch SGD: Training ImageNet in 1 hour}.
\newblock {\em arXiv preprint arXiv:1706.02677}, 2017.

\bibitem{guo2016dynamic}
Yiwen Guo, Anbang Yao, and Yurong Chen.
\newblock {Dynamic network surgery for efficient DNNs}.
\newblock In {\em NeurIPS}, pages 1379--1387, 2016.

\bibitem{han2}
Song Han, Huizi Mao, and William~J Dally.
\newblock {Deep compression: Compressing deep neural networks with pruning,
  trained quantization and Huffman coding}.
\newblock In {\em ICLR}, 2015.

\bibitem{han1}
Song Han, Jeff Pool, John Tran, and William Dally.
\newblock Learning both weights and connections for efficient neural network.
\newblock In {\em NeurIPS}, pages 1135--1143, 2015.

\bibitem{resnet}
Kaiming He, Xiangyu Zhang, Shaoqing Ren, and Jian Sun.
\newblock Deep residual learning for image recognition.
\newblock In {\em CVPR}, pages 770--778, 2016.

\bibitem{he2018amc}
Yihui He, Ji Lin, Zhijian Liu, Hanrui Wang, Li-Jia Li, and Song Han.
\newblock {AMC: AutoML for model compression and acceleration on mobile
  devices}.
\newblock In {\em ECCV}, volume 11211 of {\em LNCS}, pages 815--832, 2018.

\bibitem{he2019filter}
Yang He, Ping Liu, Ziwei Wang, Zhilan Hu, and Yi Yang.
\newblock Filter pruning via geometric median for deep convolutional neural
  networks acceleration.
\newblock In {\em CVPR}, pages 4340--4349, 2019.

\bibitem{he2017channel}
Yihui He, Xiangyu Zhang, and Jian Sun.
\newblock Channel pruning for accelerating very deep neural networks.
\newblock In {\em ICCV}, pages 1389--1397, 2017.

\bibitem{hinton2014distilling}
Geoffrey Hinton, Oriol Vinyals, and Jeff Dean.
\newblock Distilling the knowledge in a neural network.
\newblock In {\em Deep Learning and Representation Learning Workshop, NIPS},
  2014.

\bibitem{huang2018data}
Zehao Huang and Naiyan Wang.
\newblock Data-driven sparse structure selection for deep neural networks.
\newblock In {\em ECCV}, volume 11220 of {\em LNCS}, pages 317--334, 2018.

\bibitem{BN}
Sergey Ioffe and Christian Szegedy.
\newblock Batch normalization: Accelerating deep network training by reducing
  internal covariate shift.
\newblock In {\em ICML}, pages 448--456, 2015.

\bibitem{Dogs}
Aditya Khosla, Nityananda Jayadevaprakash, Bangpeng Yao, and Li Fei-Fei.
\newblock Novel dataset for fine-grained image categorization.
\newblock In {\em First Workshop on Fine-Grained Visual Categorization, CVPR},
  2011.

\bibitem{Car}
Jonathan Krause, Michael Stark, Jia Deng, and Li Fei-Fei.
\newblock 3d object representations for fine-grained categorization.
\newblock In {\em 4th IEEE Workshop on 3D Representation and Recognition,
  ICCV}, 2013.

\bibitem{CIFAR}
Alex Krizhevsky.
\newblock Learning multiple layers of features from tiny images.
\newblock Master's thesis, University of Toronto, 2009.

\bibitem{AlexNet}
Alex Krizhevsky, Ilya Sutskever, and Geoffrey~E Hinton.
\newblock {ImageNet classification with deep convolutional neural networks}.
\newblock In {\em NeurIPS}, pages 1097--1105, 2012.

\bibitem{MNIST}
Yann LeCun, L{\'e}on Bottou, Yoshua Bengio, and Patrick Haffner.
\newblock {Gradient-based learning applied to document recognition}.
\newblock {\em Proceedings of the IEEE}, 86(11):2278--2324, 1998.

\bibitem{weight_sum}
Hao Li, Asim Kadav, Igor Durdanovic, Hanan Samet, and Hans~Peter Graf.
\newblock Pruning filters for efficient convnets.
\newblock In {\em ICLR}, 2017.

\bibitem{GDP}
Shaohui Lin, Rongrong Ji, Yuchao Li, Yongjian Wu, Feiyue Huang, and Baochang
  Zhang.
\newblock Accelerating convolutional networks via global \& dynamic filter
  pruning.
\newblock In {\em IJCAI}, pages 2425--2432, 2018.

\bibitem{GAL}
Shaohui Lin, Rongrong Ji, Chenqian Yan, Baochang Zhang, Liujuan Cao, Qixiang
  Ye, Feiyue Huang, and David Doermann.
\newblock {Towards optimal structured CNN pruning via generative adversarial
  learning}.
\newblock In {\em CVPR}, pages 2790--2799, 2019.

\bibitem{liu2017learning}
Zhuang Liu, Jianguo Li, Zhiqiang Shen, Gao Huang, Shoumeng Yan, and Changshui
  Zhang.
\newblock Learning efficient convolutional networks through network slimming.
\newblock In {\em ICCV}, pages 2736--2744, 2017.

\bibitem{liu2018rethinking}
Zhuang Liu, Mingjie Sun, Tinghui Zhou, Gao Huang, and Trevor Darrell.
\newblock Rethinking the value of network pruning.
\newblock In {\em ICLR}, 2019.

\bibitem{cosine_decay}
Ilya Loshchilov and Frank Hutter.
\newblock {SGDR: Stochastic gradient descent with warm restarts}.
\newblock In {\em ICLR}, 2017.

\bibitem{luo2018autopruner}
Jian-Hao Luo and Jianxin Wu.
\newblock {AutoPruner: An end-to-end trainable filter pruning method for
  efficient deep model inference}.
\newblock {\em arXiv preprint arXiv:1805.08941}, 2018.

\bibitem{luo2017thinet}
Jian-Hao Luo, Jianxin Wu, and Weiyao Lin.
\newblock {ThiNet: A filter level pruning method for deep neural network
  compression}.
\newblock In {\em ICCV}, pages 5058--5066, 2017.

\bibitem{molchanov2019importance}
Pavlo Molchanov, Arun Mallya, Stephen Tyree, Iuri Frosio, and Jan Kautz.
\newblock Importance estimation for neural network pruning.
\newblock In {\em CVPR}, pages 11264--11272, 2019.

\bibitem{pets}
Omkar~M Parkhi, Andrea Vedaldi, Andrew Zisserman, and CV Jawahar.
\newblock Cats and dogs.
\newblock In {\em CVPR}, pages 3498--3505, 2012.

\bibitem{pytorch}
Adam Paszke, Sam Gross, Soumith Chintala, Gregory Chanan, Edward Yang, Zachary
  DeVito, Zeming Lin, Alban Desmaison, Luca Antiga, and Adam Lerer.
\newblock Automatic differentiation in pytorch.
\newblock In {\em Autodiff Workshop, NIPS}, 2017.

\bibitem{ImageNet}
Olga Russakovsky, Jia Deng, Hao Su, Jonathan Krause, Sanjeev Satheesh, Sean Ma,
  Zhiheng Huang, Andrej Karpathy, Aditya Khosla, Michael Bernstein,
  Alexander~C. Berg, and Li Fei-Fei.
\newblock {ImageNet large scale visual recognition challenge}.
\newblock {\em IJCV}, 115(3):211--252, 2015.

\bibitem{mobilenetv2}
Mark Sandler, Andrew Howard, Menglong Zhu, Andrey Zhmoginov, and Liang-Chieh
  Chen.
\newblock {MobileNetV2: Inverted residuals and linear bottlenecks}.
\newblock In {\em CVPR}, 2018.

\bibitem{VGG16}
Karen Simonyan and Andrew Zisserman.
\newblock {Very deep convolutional networks for large-scale image recognition}.
\newblock In {\em ICLR}, 2015.

\bibitem{CUB200}
Catherine Wah, Steve Branson, Peter Welinder, Pietro Perona, and Serge
  Belongie.
\newblock {The Caltech-UCSD birds-200-2011 dataset}.
\newblock Technical Report CNS-TR-2011-001, California Institute of Technology,
  2011.

\bibitem{R2D2}
Guo-Hua Wang and Jianxin Wu.
\newblock Repetitive reprediction deep decipher for semi-supervised learning.
\newblock {\em arXiv preprint arXiv:1908.04345}, 2019.

\bibitem{pencil}
Kun Yi and Jianxin Wu.
\newblock Probabilistic end-to-end noise correction for learning with noisy
  labels.
\newblock In {\em CVPR}, pages 7017--7025, 2019.

\bibitem{slimmable}
Jiahui Yu, Linjie Yang, Ning Xu, Jianchao Yang, and Thomas Huang.
\newblock Slimmable neural networks.
\newblock In {\em ICLR}, 2019.

\bibitem{mixup}
Hongyi Zhang, Moustapha Cisse, Yann~N Dauphin, and David Lopez-Paz.
\newblock mixup: Beyond empirical risk minimization.
\newblock In {\em ICLR}, 2018.

\end{thebibliography}
}

\end{document}